\documentclass{article} 
\usepackage{iclr2018_conference,times}
\usepackage{hyperref}
\usepackage{url}
\usepackage{booktabs}
\usepackage{xcolor,soul}
\hypersetup{
    colorlinks,
    linkcolor={red!50!black},
    citecolor={blue!50!black},
    urlcolor={blue!80!black}
}
\usepackage{multirow}
\usepackage{diagbox}
\usepackage{wrapfig}
\newcommand{\Bl}[1]{\textcolor{blue}{#1}}
\newcommand{\nematus}{\texttt{Nematus}} 
\newcommand{\chartochar}{\texttt{char2char}} 
\newcommand{\charcnn}{\texttt{charCNN}}
\newcommand{\meanchar}{\texttt{meanChar}}

\usepackage[colorinlistoftodos,prependcaption,textsize=small]{todonotes}

\usepackage{textcomp}

\newenvironment{myquote}%
  {\list{}{\leftmargin=0.20in\rightmargin=0.19in}\item[]}%
  {\endlist}
\newenvironment{myquotenarrow}%
  {\list{}{\leftmargin=0.10in\rightmargin=0.06in}\item[]}%
  {\endlist}

\title{Synthetic and Natural Noise Both Break\\ Neural Machine Translation}

\author{Yonatan Belinkov\thanks{Equal contribution. Ordering determined by bartender's coin: \url{https://youtu.be/BFSc2HnpYtA}}\\
Computer Science and\\
Artificial Intelligence Laboratory, \\
Massachusetts Institute of Technology\\
\texttt{belinkov@mit.edu} 
\And
Yonatan Bisk$^{*}$ \\
Paul G. Allen School\\
of Computer Science \& Engineering, \\
University of Washington\\
\texttt{ybisk@cs.washington.edu} \\
}

\iclrfinalcopy 

\begin{document}

\maketitle

\begin{abstract}
Character-based neural machine translation (NMT) models alleviate out-of-vocabulary issues, learn morphology, and move us closer to completely end-to-end translation systems.  Unfortunately, they are also very brittle and easily falter when presented with noisy data.  
In this paper, we confront NMT models with synthetic and natural sources of noise. We find that state-of-the-art models fail to translate even moderately noisy texts that humans have no trouble comprehending. We explore two approaches to increase model robustness: structure-invariant word representations and robust training on noisy texts. We find that a model based on a character convolutional neural network is able to simultaneously learn representations robust to multiple kinds of noise. 
\end{abstract}

\section{Introduction}
Humans have surprisingly robust language processing systems that 
can easily overcome typos, misspellings, and the complete omission of letters when reading \citep{rawlinson}.  A particularly extreme and comical exploitation of our robustness came years ago in the form of a popular meme:
\begin{myquote}
``Aoccdrnig to a rscheearch at Cmabrigde Uinervtisy, it deosn't mttaer in waht oredr the ltteers in a wrod are, the olny iprmoetnt tihng is taht the frist and lsat ltteer be at the rghit pclae.''
\end{myquote}

A person's ability to read this text comes as no surprise to the psychology 
literature.  \cite{saberi} found that this robustness extends to audio as well.  They experimented with playing parts of audio transcripts backwards and found that it did not affect comprehension.  \cite{slower} found that in noisier settings reading comprehension only slowed by 11\%. \cite{swapping} found that the common case of swapping letters could often go unnoticed by the reader.  The exact mechanisms and limitations of our understanding system are unknown.  There is some evidence that we rely on word shape \citep{Mayall}, that we can switch between whole word recognition and piecing together words from letters \citep{Reicher, Pelli}, and there appears to be no evidence that the first and last letter positions are required to stay constant for comprehension.\footnote{One caveat we feel is important to note is that most of the literature in psychology 
has focused on English.}

In stark contrast, neural machine translation (NMT) systems, despite their pervasive use, are immensely brittle.  
For instance, Google Translate 
produces the following unintelligible translation for a German version of the above meme:\footnote{Retrieved on February 2, 2018.} 
\begin{myquotenarrow}
``After being stubbornly defiant, it is clear to kenie Rlloe in which Reiehnfogle is advancing the boulders in a Wrot that is integral to Sahce, as the utterance and the lukewarm boorstbaen stmimt.''
\end{myquotenarrow}

While typos and noise are not new to NLP, our systems are rarely trained to explicitly address them, as we instead hope that the relevant noise will occur in the training data.

Despite these weaknesses, the move to character-based NMT 
is important.  It helps us tackle the long tailed distribution of out-of-vocabulary words in natural language, as well as reduce computation load of dealing with large word embedding matrices. NMT models based on characters and other sub-word units 
are able to extract stem and morphological information to generalize to unseen words and conjugations. They perform very well in practice on a range of languages~\citep{P16-1162,wu2016google}. In many cases, these models actually discover an impressive amount of morphological information about a language \citep{belinkov-EtAl:2017:Long}.  Unfortunately, training (and testing) on clean data makes models brittle and, arguably, unfit for broad deployment. 

\begin{figure}[h]
\centering
\begin{tabular}{cc}
\includegraphics[width=6.5cm]{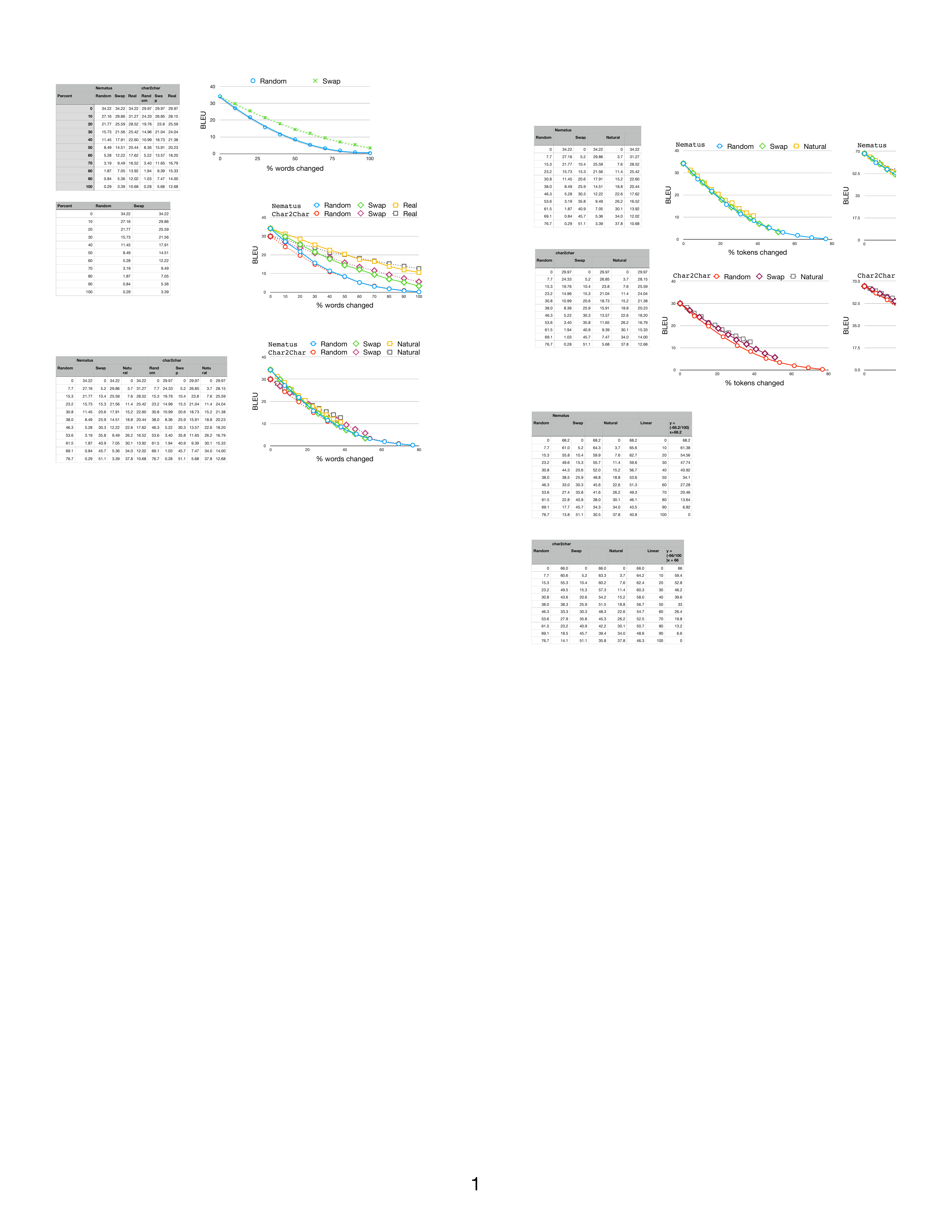} & 
\includegraphics[width=6.5cm]{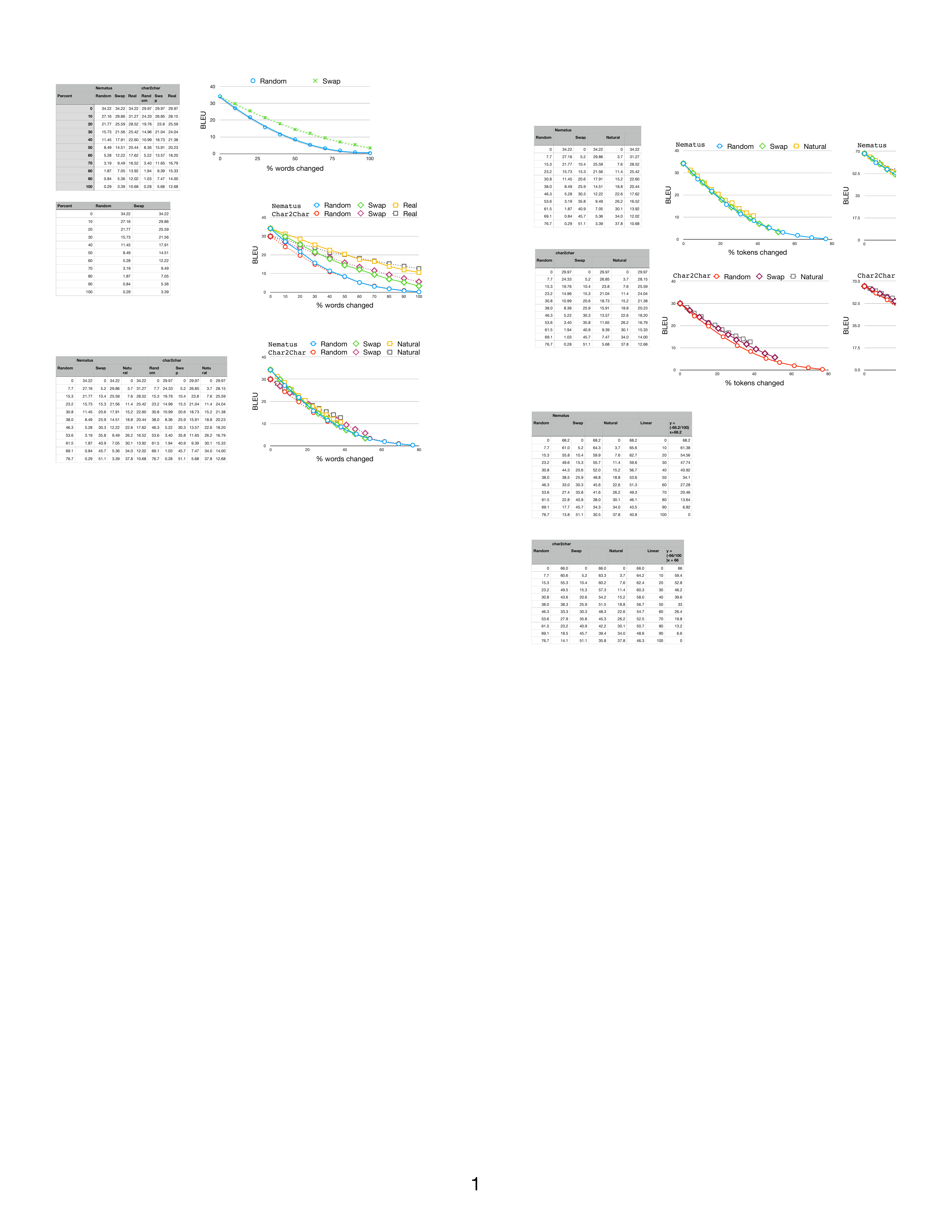}{}\\
\end{tabular}
\caption{Degradation of \texttt{Nematus}~\citep{sennrich2017} and \chartochar~\citep{lee2017} performance as noise increases.}
\label{trends}
\end{figure}

Figure \ref{trends} shows how 
the performance of two state-of-the-art NMT systems degrades when translating German to English as a function of the percent of German words modified. 
Here we show three types of noise: 1) Random permutation of the word, 2) Swapping a pair of adjacent letters,
and 3) Natural 
human errors.  We discuss these types of noise and others in depth in section \ref{sec:noise}. The important thing to note is that even small amounts of noise lead to substantial drops in performance.  

To address these trends and investigate the effects of noise on NMT, we explore two simple strategies for increasing model robustness: using structure-invariant representations and robust training on noisy data, a form of adversarial training~\citep{szegedy2013intriguing,goodfellow2014explaining}. We find that a  character CNN representation trained on an ensemble of noise types is robust to all kinds of noise. We shed some light on the model ability to learn robust representations to multiple types of noise, and point to remaining difficulties in handling natural noise. Our goal is two fold:  1) initiate a conversation on robust training and modeling techniques 
in NMT, 
and 2) promote the creation of better and more linguistically accurate artificial noise to be applied to new languages 
and tasks.

\vspace{-4pt}
\section{Adversarial Examples}
\vspace{-5pt}

The growing literature on adversarial examples has demonstrated how dangerous it can be to use brittle machine learning systems so pervasively in the real world~\citep{Biggio:2012:PAA:3042573.3042761,szegedy2013intriguing,goodfellow2014explaining,Mei:2015:UMT:2886521.2886721}.
Small changes to the input can lead to dramatic failures of deep learning models~\citep{szegedy2013intriguing,goodfellow2014explaining}. In the machine vision field, 
changes to the input image that are indistinguishable by humans can lead to 
misclassification. This leads to potential for malicious attacks using 
adversarial examples. 
An important distinction is often drawn between white-box attacks, where adversarial examples are generated with access to the  
model parameters, and black-box attacks, where examples are generated without such access~\citep{papernot2016transferability,Papernot:2017:PBA:3052973.3053009,8014906,liu2016delving}. 

While more common in the vision domain, recent work has 
started exploring adversarial examples for NLP.  A few white-box attacks have employed the fast gradient sign method~\citep{goodfellow2014explaining} or other techniques to find important text edit operations~\citep{papernot2016crafting,samanta2017towards,liang2017deep,ebrahimi2017hotflip}.
Others have considered black-box adversarial examples for text classification~\citep{gao2018black} or NLP evaluation~\citep{jia-liang:2017:EMNLP2017}. 
\cite{heigold2017robust} evaluated character-based models on several types of noise in morphological tagging and MT, and observed similar trends to our findings. Finally, \cite{DBLP:conf/aaai/SakaguchiDPD17} designed a character-level recurrent neural network that can better handle the particular kind of noise present in the meme mentioned above by modeling spelling correction. 
Here we devise simple methods for generating adversarial examples for NMT. We do not assume any access to the NMT models' gradients, instead relying on synthetic and naturally occurring language errors to generate noise.

The other side of the coin is to improve 
models' robustness to adversarial examples~\citep{Globerson:2006:NTT:1143844.1143889,Cretu:2008:COD:1397759.1398062,Rubinstein:2009:AUD:1644893.1644895,Chan2017}. Adversarial training -- including adversarial examples in the training data -- can improve a model's ability to cope with such examples at test time~\citep{szegedy2013intriguing,goodfellow2014explaining}. This kind of defense is sensitive to the type of adversarial examples seen in training, but can be made more robust by \textit{ensemble adversarial training} -- training on examples transfered from multiple pre-trained models~\citep{tramer2017ensemble}.
We explore ensemble training by combining multiple types of noise at training time, and observe similar increased robustness in the machine translation scenario. 

Training on and for adversarial noise is an important extension of earlier work 
on creating robustness in neural networks by incorporating noise to a network's representations, data, or gradients.  Training with noise can provide a form of regularization \citep{Bishop:1995} and ensure the model is exposed to samples outside the training distribution \citep{Matsuoka:1992}.

\vspace{-5pt}
\section{MT Systems}
\vspace{-5pt}
 The rise of end-to-end models in neural machine translation has led to recent interest in understanding how these models operate. Several studies investigated the ability of such models to learn linguistic properties at morphological~\citep{vylomova2016word,belinkov-EtAl:2017:Long,dalvi:2017:IJCNLP}, syntactic~\citep{shi-padhi-knight:2016:EMNLP2016,E17-2060}, and semantic levels~\citep{belinkov:2017:IJCNLP}. The use of characters or other sub-word units emerges as an important component in these models. Our work complements previous studies by presenting such NMT systems with noisy examples and exploring methods for increasing their robustness.

We experiment with three different NMT systems with access to character information at different levels.
First, we use the fully character-level model of~\cite{lee2017}. This is a sequence-to-sequence model with attention~\citep{sutskever2014sequence,bahdanau2014neural} that is trained on characters to characters (\texttt{char2char}). It has a complex encoder with convolutional, highway, and recurrent layers, and a standard recurrent decoder. See~\cite{lee2017} for architecture details. 
This model was shown to have excellent performance on the German$\rightarrow$English and Czech$\rightarrow$English language pairs. We use the pre-trained 
German/Czech$\rightarrow$English models.  

Second, we use \texttt{Nematus}~\citep{sennrich2017}, a popular NMT toolkit that was used in top-performing contributions in shared MT tasks in 
WMT~\citep{sennrich-haddow-birch:2016:WMT} and IWSLT~\citep{junczys2016university}. It is another sequence-to-sequence model with several architecture modifications, especially operating on sub-word units using byte-pair encoding (BPE)~\citep{P16-1162}. We experimented with both their single best and ensemble BPE models, but saw no significant difference in their performance under noise, 
so we report results with their single best WMT models for German/Czech$\rightarrow$English.   

Finally, we train an attentional sequence-to-sequence model with a word representation based on a character convolutional neural network (\texttt{charCNN}). This model retains the notion of a word but learns a character-dependent representation of words.  It was shown to perform well on morphologically-rich languages~\citep{kim2015character,belinkov-glass:2016:SeMaT,
costajussa-fonollosa:2016:P16-2,sajjad:2017:ACL}, 
 thanks to its ability to learn morphologically-informative representations~\citep{belinkov-EtAl:2017:Long}. 
The \charcnn~model has two long short-term memory
~\citep{hochreiter1997long} layers in the encoder and
decoder.  
A CNN over characters in each word replaces the word embeddings on the encoder side (for simplicity, the decoder
is word-based).
We use 1000 filters with a width of 6 characters. The character embedding size is set to 25. The convolutions are followed by Tanh 
and max-pooling over the length of the word~\citep{kim2015character}. 
We train \charcnn~with the
implementation in~\cite{kim2016}; all other settings are kept to 
default values.

\vspace{-5pt}
\section{Data} 
\vspace{-5pt}
\subsection{MT Data}
We use the TED talks parallel corpus prepared for IWSLT 2016~\citep{cettoloEtAl:EAMT2012} for testing all of the NMT systems, as well as for training the \charcnn~models. We follow the official training/development/test splits. All texts are tokenized with the Moses tokenizer. 
Table~\ref{tab:data} summarizes statistics on the TED talks corpus. 

\subsection{Noise: Natural and Artificial}
\label{sec:noise}

\begin{table}[t]
\centering
\caption{Statistics for the source-side of French/German/Czech$\rightarrow$English parallel corpora.} 
\label{tab:data}
\begin{tabular}{l@{\hspace{30pt}}r@{\hspace{5pt}}r@{\hspace{5pt}}r@{\hspace{30pt}}r@{\hspace{5pt}}r@{\hspace{5pt}}r@{\hspace{30pt}}r@{\hspace{5pt}}r@{\hspace{5pt}}r}
\toprule
& \multicolumn{3}{c@{\hspace{30pt}}}{French} & \multicolumn{3}{c@{\hspace{30pt}}}{German} & \multicolumn{3}{c}{Czech} \\
& Train & Dev & Test & Train & Dev & Test & Train & Dev & Test\\
\midrule
Sentences & 235K & 2.5K & 0.8K & 210K & 2.5K & 1.4K & 122K & 20K & 1K \\
Words & 5.2M & 55K & 16K & 4M & 50K & 26K & 2.1M & 35K & 15K \\
\bottomrule
\end{tabular}
\vspace{-10pt}
\end{table}

We insert noise into the source-side of the parallel MT data by utilizing naturally occurring errors and generating synthetic ones. 
In order to facilitate future work on noise in NMT, we release code and data for generating the noise used in our experiments.\footnote{\url{https://github.com/ybisk/charNMT-noise}} 

\subsubsection{Natural noise}
Since we do not have access to a parallel corpus with natural noise, we instead harvest naturally occurring errors (typos, misspellings, etc.) from available corpora of edits to build a look-up table of possible lexical replacements.  In this work, we restrict 
ourselves to 
single word replacements, but several of the corpora below also provide access to phrase replacements. 

\paragraph{French} \cite{max10wicopaco} 
collected Wikipedia edit histories to form the Wikipedia Correction and Paraphrase Corpus (WiCoPaCo).  
They found the bulk of edits were due to incorrect diacritics, choosing the wrong homophone, and incorrect grammatical conjugation.
\paragraph{German} Our German data combines two projects: RWSE Wikipedia Revision Dataset \citep{zesch:2012:EACL2012} and The MERLIN corpus of language learners \citep{MERLIN}. 
These corpora were created to measure spelling difficulty and test models of contextual fitness. Unfortunately, the datasets are quite small so we have combined them here.  

\paragraph{Czech} Our Czech errors come from manually annotated essays written by non-native speakers \citep{CzeSL}. 
Here, the authors found an incredibly diverse set of errors, and therefore phenomena of interest: capitalization, incorrectly replacing voiced and voiceless consonants (e.g.~z/s, g/k), missing palatalization (mat\v{k}e/matce), error in valence, pronominal reference, inflection, colloquial forms, and so forth.  Their analysis gives us the best insight into how difficult it would be to synthetically generate truly natural errors. We found similarly rich errors in  German  (Section~\ref{sec:analysis-noise}). 

\begin{table}[t]
\centering
\caption{Average number of available edits per word in natural noise datasets and the corresponding token recall of those edits on the training and test splits.}
\label{natural}
\begin{tabular}{c@{\hspace{5pt}}c@{\hspace{5pt}}c@{\hspace{5pt}}c@{\hspace{30pt}}c@{\hspace{5pt}}c@{\hspace{5pt}}c@{\hspace{5pt}}c@{\hspace{30pt}}c@{\hspace{5pt}}c@{\hspace{5pt}}c@{\hspace{5pt}}c}
\toprule
   \multicolumn{4}{c@{\hspace{25pt}}}{French}             & \multicolumn{4}{c@{\hspace{25pt}}}{German}             & \multicolumn{4}{c}{Czech}             \\
   Words & Errors & Train & Test        & Words & Errors & Train & Test        & Words & Errors & Train & Test        \\
\midrule
  65,156 & 2.7 & 40\%    & 41\%          & 1,344  & 2.5 & 37\%    & 40\%          & 6,036  & 2.6 & 46\%    & 51\% \\
\bottomrule
\end{tabular}
\vspace{-10pt}
\end{table}

We insert these errors into the source-side of the parallel data by replacing every word in the corpus with an error if one exists in our dataset.  When there is more than one possible replacement to choose we sample uniformly.  Words for which there is no error are kept as is. Table \ref{natural} shows the number of words for which we were able to collect errors in each language, and the average number of errors per word.  Despite the small size of the German and Czech datasets, we are able to replace up to half of the words in the corpus with errors.  Due to the small size of the German and Czech datasets these percentages decrease for longer words ($>4$ characters) to 25\% and 32\%, respectively.

\subsubsection{Synthetic noise}
In addition to naturally collected sources of error, we also experiment with four types of synthetic noise: Swap, Middle Random, Fully Random, and Keyboard Typo.
\paragraph{Swap : \texttt{Swap}} The simplest source of noise is swapping two letters (e.g. {\it noise}$\rightarrow${\it nosie}).  This is common when typing quickly and is easily implemented.  We perform one swap per word, but do not alter the first or last letters.  For this reason, this noise is only applied to words of length $\geq 4$. 
\vspace{-3pt}
\paragraph{Middle Random : \texttt{Mid}} Following the claims of the previously discussed meme, we randomize the order of all the letters in a word except for the first and last ({\it noise}$\rightarrow${\it nisoe}).  Again, by necessity, this means we do not alter words shorter than four characters.
\vspace{-3pt}
\paragraph{Fully Random : \texttt{Rand}} As we are unaware of any strong results on the importance of the first and last letters we also include completely randomized words ({\it noise}$\rightarrow${\it iones}).  This is a particularly extreme case, but we include it for completeness.  This type of noise is applied to all words.
\vspace{-3pt}
\paragraph{Keyboard Typo : \texttt{Key}} Finally, using the traditional keyboards for our languages, we randomly replace one letter in each word with an adjacent key ({\it noise}$\rightarrow${\it noide}).  This type of error should be much easier than the random settings as most of the word is left intact, but does introduce a completely new character which will often break the templates a system has learned to rely on.

\section{Failures to Translate Noisy Texts}
Table~\ref{tab:res-normal-training} shows BLEU scores of models trained on clean (Vanilla) texts and tested on clean and noisy texts. All models suffer a significant drop in BLEU when evaluated on noisy texts. This is true for both natural noise and all kinds of synthetic noise. The more noise in the text, the worse the translation quality, with random scrambling producing the lowest BLEU scores. 

\begin{table}[t]
\centering
\caption{The effect of Natural (\texttt{Nat}) and synthetic noise (Swap \texttt{swap}, Middle Random \texttt{Mid}, Fully Random \texttt{Rand}, and Keyboard Typo \texttt{Key}) on models trained on clean (Vanilla) texts. }
\label{tab:res-normal-training}
\begin{tabular}{ll@{\hspace{17pt}}r@{\hspace{17pt}}r@{\hspace{7pt}}r@{\hspace{7pt}}r@{\hspace{7pt}}r@{\hspace{17pt}}r}
\toprule
                         &             &          &\multicolumn{4}{c@{\hspace{15pt}}}{Synthetic} & \\
                         &             & Vanilla  & \texttt{Swap} & \texttt{Mid} & \texttt{Rand} & \texttt{Key} & \texttt{Nat}\\
\midrule
\multirow{1}{*}{French}  & \charcnn    & 42.54 & 10.52& 9.71 & 1.71 & 8.26 & 17.42 \\
\midrule                                                                          
\multirow{3}{*}{German}  & \charcnn    & 34.79 & 9.25 & 8.37 & 1.02 & 6.40 & 14.02 \\
                         & \chartochar & 29.97 & 5.68 & 5.46 & 0.28 & 2.96 & 12.68 \\
                         & \nematus    & 34.22 & 3.39 & 5.16 & 0.29 & 0.61 & 10.68 \\
\midrule                                                                          
\multirow{3}{*}{Czech}   & \charcnn    & 25.99 & 6.56 & 6.67 & 1.50 & 7.13 & 10.20 \\
                         & \chartochar & 25.71 & 3.90 & 4.24 & 0.25 & 2.88 & 11.42 \\
                         & \nematus    & 29.65 & 2.94 & 4.09 & 0.66 & 1.41 & 11.88 \\
\bottomrule
\end{tabular}
\vspace{-10pt}
\end{table}

\begin{table}[t]
\centering
\caption{An example noisy text with human and machine translations.}
\label{tab:example}
\begin{small}
\begin{tabular}{@{}p{1.6cm}p{11.9cm}@{}}
\toprule
Input & Luat eienr Stduie der Cambrdige Unievrstiät speilt es kenie Rlloe in welcehr Reiehnfogle die Buhcstbaen in eniem Wrot vorkmomen, die eingzie whctige Sahce ist, dsas der ertse und der lettze Buhcstbaen stmimt . \\[5pt]
Human & According to a study from Cambridge university, it doesn't matter which order letters in a word are, the only important thing is that the first and the last letter appear in their correct place. \\
\midrule
\chartochar & Cambridge Universtätte is one of the most important features of the Cambridge Universtätten , which is one of the most important features of the Cambridge Universtätten . \\ 
\nematus & Luat eienr Stduie der Cambrant Unievrstilt splashed it kenie Rlloe in welcehr Reiehnfogle the Buhcstbaen in eniem Wred vorkmomen, die eingzie whcene Sahce ist, DSAs der ertse und der lettze Buhcstbaen stmimt .\\
\charcnn & According to the $<$unk$>$ of the Cambridge University , it 's a little bit of crude oil in a little bit of recycling , which is a little bit of a cool cap , which is a little bit of a strong cap , that the fat and the $<$unk$>$ bites is consistent . \\
\bottomrule
\end{tabular}
\end{small}
\vspace{-10pt}
\end{table}

The degradation in translation quality is especially severe in light of  humans' ability 
to understand noisy texts. To illustrate this, consider the  noisy text in Table~\ref{tab:example}. 
Humans are quite good at understanding such scrambled texts in a variety of languages.\footnote{Matt Davis has a wide collection of translations of this text in multiple languages:\\
\url{https://www.mrc-cbu.cam.ac.uk/personal/matt.davis/Cmabrigde/}.} 
We also verified this by obtaining a translation from a German native-speaker, unfamiliar with the meme. As shown in the table, the speaker had no trouble understanding and translating the sentence properly. In contrast, the state-of-the-art systems (\chartochar~and \nematus) fail on this text. 

One natural question is if robust spell checkers trained on human errors are sufficient to address this performance gap.  To test this, we ran texts with and without natural errors through Google Translate.  We then used Google's spell-checkers to correct the documents.  We simply accepted the first suggestion for every detected mistake detected, and report results in Table \ref{tab:google}.

\begin{table}[t]
\centering
\caption{Google Translate's performance with natural errors and the gains from using spell checking.}
\label{tab:google}
\begin{tabular}{c@{\hspace{0.7em}}c@{\hspace{0.7em}}c @{\hspace{30pt}} c@{\hspace{0.7em}}c@{\hspace{0.7em}}c @{\hspace{30pt}} c@{\hspace{0.7em}}c@{\hspace{0.7em}}c}
\toprule
		& French 		   & 		  &			& German 		   &          &         &  Czech   & \\
Vanilla & \texttt{Nat} & Spelling & Vanilla & \texttt{Nat} & Spelling & Vanilla & \texttt{Nat} & Spelling \\
\midrule
43.3    & 16.7             &  21.4         &  38.7   &  18.6            & 25.0          & 26.5    &    12.3   & 11.2 \\
\bottomrule
\end{tabular}
\vspace{-10pt}
\end{table}

We found that in French and German, there was often only a single predicted correction and this corresponds to roughly +5 or more in BLEU.  In Czech, however, there was often a large list of possible conjugations and changes, likely indicating that a rich grammatical model would be necessary to predict the correction.  It is also important to note the substantial drops from vanilla text even with spell check. This suggests that natural noise cannot be easily addressed by existing tools.

\section{Dealing with Noise}

\subsection{Structure Invariant Representations}
The three NMT models are all sensitive to word structure. The \chartochar~and \charcnn~ models both have convolutional layers on character sequences, designed to capture character n-grams. The model in \nematus~is based on sub-word units obtained with BPE. It thus relies on character order within and across sub-word units. All these models are therefore sensitive to types of noise generated by character scrambling (\texttt{Swap}, \texttt{Mid}, and \texttt{Rand}). 
Can we improve model robustness by adding invariance to these kinds of noise?  
Perhaps the simplest such model is to take the average character embedding as a word representation. This model, referred to as \meanchar, first generates a word representation by averaging character embeddings, and then proceeds with a word-level encoder similar to the \charcnn~model. 
The \meanchar~model is by definition insensitive to scrambling, although it is still sensitive to other kinds of noise (\texttt{Key} and \texttt{Nat}). 

\begin{table}[t]
\centering
\caption{Results of \meanchar~models trained and tested on different noise conditions: Scrambled (Scr), Keyboard Typo (\texttt{Key}), and Natural (\texttt{Nat}).}
\label{tab:res-robust-training-mean}
\begin{tabular}{l|rrr|rrr|rrr}
\toprule
& \multicolumn{3}{c|}{French} & \multicolumn{3}{c|}{German} & \multicolumn{3}{c}{Czech} \\
\backslashbox{Train}{Test} & Scr & \texttt{Key}  & \texttt{Nat}      & Scr & \texttt{Key}  & \texttt{Nat}      & Scr & \texttt{Key}  & \texttt{Nat}     \\
\midrule
Vanilla         & 34.26 &  4.27 & 12.58 & 27.53 & 3.34 & 9.41  & 3.73 & 2.06 & 3.25 \\
\midrule\midrule
\texttt{Key}              & 31.88 & 29.75 & 13.16 & 10.04 & 8.84 & 4.45  & 2.03 & 1.9  & 1.42 \\
\texttt{Nat}             & 26.94 & 5.30  & 27.49 & 15.65 & 3.06 & 26.26 &   1.66   &   1.52   &   1.58   \\
\texttt{Rand} + \texttt{Key}        & 13.60 & 11.09 & 6.12  & 26.59 & 22.41& 11.07 & 9.97 & 7.48 & 4.21 \\ 
\texttt{Rand} + \texttt{Nat}       & 28.28 & 5.10  & 20.40 & 13.87 & 3.73 & 12.74 &   4.89   &  2.82    &  3.42    \\
\texttt{Key} +  \texttt{Nat}        & 31.30 & 26.94 & 24.24 & 6.62  & 5.41 & 5.75  &   1.62   &  1.68    &  1.58    \\
\texttt{Rand} + \texttt{Key} + \texttt{Nat} & 3.10  & 3.28  & 2.76  & 8.02  & 5.79 & 6.36  &   1.73   &   1.74   &  1.66    \\
\bottomrule
\end{tabular}
\vspace{-10pt}
\end{table}

Table~\ref{tab:res-robust-training-mean} (first row) shows the results of \meanchar~models trained on vanilla texts 
and tested on noisy texts (the results on vanilla texts are by definition equal to those on scrambled texts). 
Overall, the average character embedding proves to be a pretty good representation for translating scrambled texts: while performance drops by about 7 BLEU points below \charcnn~on vanilla French and German, it is much better than \charcnn's performance on scrambled texts (compare to Table~\ref{tab:res-normal-training}). The results of \meanchar~on Czech are much worse, possibly due to its more complex morphology. 
However, the \meanchar~model performance degrades quickly on other kinds of noise as the model trained on vanilla texts was not designed to handle \texttt{Nat} and \texttt{Key} types of noise.

\begin{table}[t]
\caption{Results of \charcnn~models trained and tested on different noise conditions.}
\label{tab:res-robust-training}
\centering
\begin{tabular}{l|l|r|rrrrr|r}
\toprule
& \backslashbox{Train}{Test} & Vanilla & \texttt{Swap} & \texttt{Mid} & \texttt{Rand} & \texttt{Key}  & \texttt{Nat}  & Ave\\
\midrule
\multirow{6}{*}{French} 
 & \texttt{Swap}          & 39.01 &\bf 42.56 & 33.64 & 2.72 & 4.85 & 16.43  & 23.20 \\
 & \texttt{Mid}        &\bf 42.46 & 42.19 &\bf 42.17 & 3.36 & 6.20 & 18.22  & 25.77 \\
 & \texttt{Rand}          & 39.53 & 39.46 & 39.13 &\bf 39.73 & 3.11 & 16.63  & 29.60 \\
 & \texttt{Key}           & 38.49 & 10.56 & 8.69 & 1.08 &\bf 38.88 & 16.86  & 19.10 \\
 & \texttt{Nat}          & 28.77 & 12.45 & 8.39 & 1.03 & 6.61 &\bf 36.00  & 15.54\\
 & \texttt{Rand} + \texttt{Key}  & 39.23 & 38.85 & 38.89 &\Bl{39.13}&\Bl{38.22}& 18.71  & 35.51\\
 & \texttt{Rand} + \texttt{Nat} & 36.86 & 38.95 & 38.44 &\Bl{38.63}& 6.67 &\Bl{33.89} & 32.24\\
 & \texttt{Key}  + \texttt{Nat}    & 38.47 & 17.33 & 10.54 & 1.52 &\Bl{38.62}&\Bl{34.66} & 23.52\\
 & \texttt{Rand} + \texttt{Key} + \texttt{Nat} & 36.97 & 36.92 & 36.65 &\Bl{36.64}&\Bl{35.25}&\Bl{31.77} & \bf \Bl{35.70}\\
\midrule \midrule
\multirow{4}{*}{German} 
 & \texttt{Swap}    & 32.66 &\bf 34.76 & 29.03 & 2.19 & 4.78 & 13.37  & 19.47\\
 & \texttt{Mid}     &\bf 34.32 & 34.26 &\bf 34.27 & 3.50 & 5.08 & 14.43  & 20.98\\
 & \texttt{Rand}    & 33.65 & 33.44 & \bf 33.75 & 33.56 & 3.00 & 14.47  & 25.31 \\
 & \texttt{Key}     & 32.87 & 10.13 & 8.39 & 1.16 &\bf 33.28 & 13.88  & 16.62\\
 & \texttt{Nat}     & 25.79 & 8.20 & 5.73 & 0.93 & 4.80 &\bf 34.59  & 13.34\\
 & \texttt{Rand} + \texttt{Key} & 32.03 & 31.57 & 31.32 &\Bl{31.58} &\Bl{31.23} & 15.59  & 28.89\\ 
 & \texttt{Rand} + \texttt{Nat} & 32.37 & 32.40 & 31.91 &\Bl{32.11} & 4.77 &\Bl{33.00}  & 27.76\\
 & \texttt{Key}  + \texttt{Nat} & 30.39 & 13.51 & 8.99 & 1.53 &\Bl{32.23} &\Bl{33.46}  & 20.02\\
 & \texttt{Rand} + \texttt{Key} + \texttt{Nat} & 31.29 & 30.93 & 30.54 &\Bl{30.04} &\Bl{29.81} &\Bl{31.60}  & \bf \Bl{30.70}\\
\midrule \midrule
\multirow{4}{*}{Czech} 
 & \texttt{Swap}   &\bf 24.22 &\bf 24.90 & 18.72 & 2.72 & 6.00 & 9.03  & 14.27 \\
 & \texttt{Mid}    & 23.81 & \bf 24.52 & 24.08 & 3.96 & 6.34 & 9.54  & 15.38 \\
 & \texttt{Rand}   & 23.44 & 23.31 & 23.24 &\bf 23.47 & 3.70 & 8.10  & 17.54 \\
 & \texttt{Key}    & 23.15 & 7.06 & 6.04 & 1.56 &\bf 22.80 & 10.16  & 11.80  \\
 & \texttt{Nat}    & 18.04 & 5.36 & 4.48 & 1.47 & 6.71 & \bf 21.64 & 9.62 \\
 & \texttt{Rand} + \texttt{Key} & 21.46 & 20.81 & 20.90 & \Bl{20.59} & \Bl{19.48} & 8.72  & 18.66 \\
 & \texttt{Rand} + \texttt{Nat} & 20.59 & 21.56 & 20.49 & \Bl{20.53} & 5.89 & \Bl{18.39} & 17.91 \\
 & \texttt{Key}  + \texttt{Nat} & 19.55 & 6.59 & 5.72 & 1.40 & \Bl{21.31} & \Bl{19.54} & 12.35 \\
 & \texttt{Rand} + \texttt{Key} + \texttt{Nat} & 21.30 & 21.33 & 20.38 & \Bl{19.94} & \Bl{19.25} & \Bl{18.38} &\bf  \Bl{20.10} \\
\bottomrule
\end{tabular}
\vspace{-5pt}
\end{table}

\subsection{Black-Box adversarial training}

To increase model robustness we follow a black-box adversarial training scenario, where the model is presented with adversarial examples that are generated without direct access to the model~\citep{papernot2016transferability,Papernot:2017:PBA:3052973.3053009,liu2016delving,8014906,jia-liang:2017:EMNLP2017}. 
We replace the original training set with a noisy training set,
 where noise is introduced according to the description in Section~\ref{sec:noise}. 
 The noisy training set has exactly the same number of sentences and words as the training set. We have one fixed noisy training set per each noise type.\footnote{When replacing words in the input, we inevitably make some of the same replacements on both the training and test sets.  We verify this does not decrease the percent of unseen words in testing. Conversely, we found it increases for all synthetic noise types and is similar for the vanilla and natural noise conditions.}

As shown in Table~\ref{tab:res-robust-training-mean} (second block), training on noisy text can lead to improved performance. The \meanchar~models trained on \texttt{Key} perform well on \texttt{Key} in French, but not in the other languages. The models trained on \texttt{Nat} perform well in French and German, but not in Czech. 
Overall, training the \meanchar~model on noisy text does not appear to consistently increase its robustness to different kinds of noise. 
The \meanchar~model however was not expected to perform well on non-scrambling types of noise. Next we test whether the more complicated \charcnn~model is more robust to different kinds of noise, by training on noisy texts. The results are shown in Table~\ref{tab:res-robust-training}. 

In general, \charcnn~models that are trained on a specific kind of noise perform well on the same kind of noise at test time (results in {\bf bold}). All models also maintain a fairly good quality on vanilla texts.
The robust training is sensitive to the kind of noise. Among the scrambling methods (\texttt{Swap}/\texttt{Mid}/\texttt{Rand}), more noise helps in training: models trained on \texttt{random} noise can still translate \texttt{Swap}/\texttt{Mid} noise, but not vice versa. The three broad classes of noise (scrambling, \texttt{Key}, \texttt{Nat}) are not mutually-beneficial. Models trained on one do not perform well on the others. In particular, only models trained on natural noise can reasonably translate natural noise at test time. 
We find this result indicates an important difference between computational models and human performance, since humans can decipher random letter orderings without explicit training of this form.

Next, we test whether we can increase training robustness by exposing the model to multiple types of noise during training. Our motivation is to see if models can perform well on more than one kind of noise. We therefore mix up to three kinds of noise by sampling a noise method uniformly at random for each sentence. We then train a model on the mixed noisy training set and test it on both vanilla and (unmixed) noisy versions of the test set. We find that models trained on mixed noise are slightly worse than models trained on unmixed noise. However, the models trained on mixed noise are robust to the specific types of noise they were trained on. In particular, the model trained on a mix of \texttt{Rand}, \texttt{Key}, and \texttt{Nat} noise is robust to all noise kinds. Even though it is not the best on any one kind of noise, it achieves the best result on average. 

This model is also able to translate the scrambled meme reasonably well: 
\begin{myquote}
``According to a study of Cambridge University, it doesn't matter which technology in a word is going to get the letters in a word that is the only important thing for the first and last letter.''
\end{myquote}

\section{Analysis}
\subsection{Learning multiple kinds of noise in \charcnn}
The \charcnn~model was able to perform well on all kinds of noise by training on a mix of noise types. In particular, it performed well on scrambled characters even though its convolutions should be sensitive to the character order, as opposed to \meanchar~which is by definition invariant to character order. How then can \charcnn~learn to be robust to multiple kinds of noise at the same time? 
We speculate that different convolutional filters learn to be robust to different kinds of noise. A convolutional filter can in principle capture a mean (or sum) operation by employing equal or close to equal weights. 

To test this, we analyze the weights learned by \charcnn~models trained under four conditions: three models trained each on completely scrambled words (\texttt{Rand}), keyboard typos (\texttt{Key}), and natural human errors (\texttt{Nat}), as well as an ensemble model trained on a mix of \texttt{Rand}+\texttt{Key}+\texttt{Nat} kinds of noise. 
For each model, 
we compute the variance across the filter width (6 characters)  for each one of the 1000 filters and for each one out of 25 character embedding dimensions. 
Intuitively, this variance captures how much a particular filter learns a uniform vs.\ non-uniform combination of characters. 
Then we average the variances across the 1000 filters. This yields 25 averaged variances, one for each character embedding dimension. 
Low average variance means that different filters tend to learn similar behaviors, while high average variance means that they learn different patterns. 

\begin{figure}[t]
\centering
\includegraphics[width=1.0\textwidth]{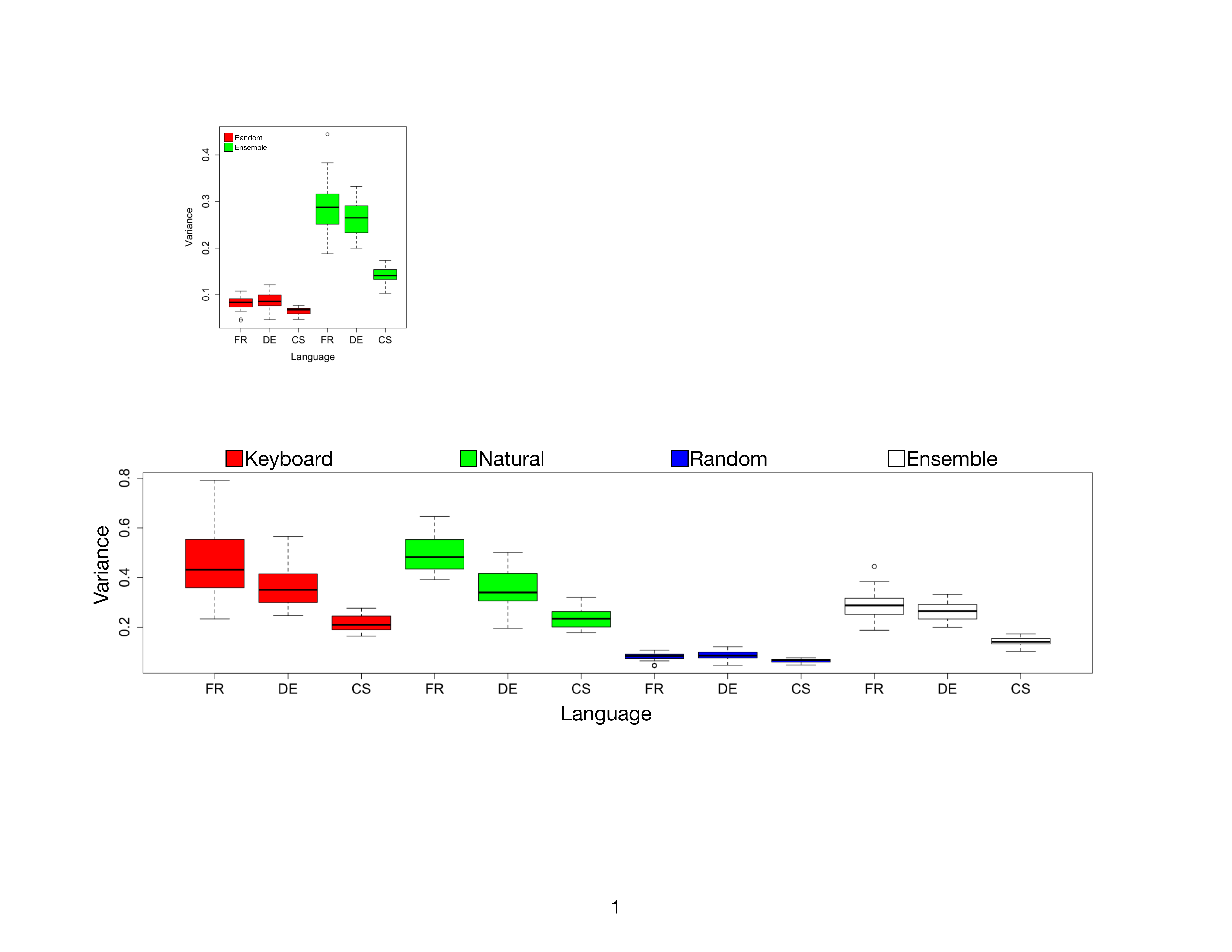}
\caption{Variances of \charcnn~weights when trained on only \texttt{Key}, \texttt{Natural}, \texttt{Random} noise and on a mix of all three are shown in red, green, blue, and white, respectively}
\label{fig:charCNN-weights}
\end{figure}

Figure~\ref{fig:charCNN-weights} shows a box plot of these averages for our three languages and four training conditions. Clearly, the variances of the weights learned by the \texttt{Rand} model are much smaller than those of the weights learned by any other setting. This makes sense as with random scrambling there are no patterns to detect in the data, so filters resort to close to uniform weights.
In contrast, the \texttt{Key} and \texttt{Nat} settings introduce a large set of new patterns for the CNNs to try and learn, leading to high variances. Finally, the ensemble model trained on mixed noise appears to be in the middle as it tries to capture both the uniform relationships of \texttt{Rand} and the more diverse patterns 
of \texttt{Nat + Key}. 

Moreover, the variance of variances (size of the box) is smallest in the \texttt{Rand} setting, larger in the mixed noise model, and largest in \texttt{Key} and \texttt{Nat}. This indicates that filters for different character embedding dimensions are more different from one another in \texttt{Key} and \texttt{Nat} models. In contrast, in the \texttt{Rand} model, the variance of variances is close to zero, indicating that in all character embedding dimensions the learned weights are of small variance; they do similar things, that is, the model learned to reproduce a representation similar to the \meanchar~model. The ensemble model again seems to find a balance between \texttt{Rand} and \texttt{Key}/\texttt{Nat}. 

\subsection{Richness of Natural Noise} \label{sec:analysis-noise}
Natural noise appears to be very different from synthetic noise. None of the models that were trained only on synthetic noise were able to perform well on natural noise. We manually analyzed a small sample (\texttildelow40 examples) of natural noise from the German 
dataset. We found that the most common sources of noise are phonetic or phonological phenomena in the language (34\%) and character omissions (32\%). The rest are incorrect 
morphological conjugations of verbs, key swaps, character insertions, orthographic variants, and other errors. Table~\ref{tab:real-noise-examples} shows examples of these kinds of noise. 

The most common types of natural noise -- phonological and omissions -- are not directly captured by our synthetic noise generation, and demonstrate that good synthetic errors will likely require more explicit phonemic and linguistic knowledge.  This discrepancy helps explain why the models trained on synthetic noise were not particularly successful in translating natural noise.

\begin{table}[h]
\caption{Examples of natural noise from the German errors dataset.}
\label{tab:real-noise-examples}
\centering
\begin{small}
\begin{tabular}{@{}l|p{11.4cm}@{}}
\toprule
Error type & Examples  \\ 
\midrule
Phonetic & Tut/Tud (devoicing of final stops), sieht/zieht (s = /z/ before vowel), Trotzdem/Trozdem \mbox{(tz = /z/)}, gekriegt/gekrigt (vowel length), Nat\"{u}rlich/Naturlich/N\"{a}turlich (diacritics) \\ 
Omission & erfahren/erfaren, Babysitter/Babysiter, selbst/sebst, Hausschuhe/Hausschue  \\
Morphological & wohnt/wonnen, fortsetzt/forzusetzen, w\"{u}nsche/w\"{u}nchen  \\
Key swap & Eltern/Eltren, Deine/Diene, nichts/nichst, Bahn/Bhan  \\
Other & Agglomerationen/Agromelationen (omission + letter swap), Hausaufgabe/Hausausgabe, 
 Thema/Temer, Detailhandelsfachfrau/Deitellhandfachfrau  \\ 
\bottomrule
\end{tabular}
\end{small}
\end{table}

\section{Conclusion}
In this work, we have shown that character-based NMT models are extremely brittle and tend to break when presented with both natural and synthetic kinds of noise. We investigated methods for increasing their robustness by using a structure-invariant word representation and by ensemble training on adversarial examples of different kinds. We found that a character-based CNN can learn to address multiple types of errors that are seen in training. However, we observed rich characteristics of natural human errors that cannot be easily captured by existing models. 
Future work might investigate using phonetic and syntactic structure to generate more realistic synthetic noise.

We believe that more work is necessary in order to immune NMT models against natural noise. As corpora with natural noise are limited, another approach to future work is to design better NMT architectures that would be robust to noise without seeing it in the training data. 
New psychology results on how humans cope with natural noise might point to possible solutions to this problem.

\section*{Acknowledgements}
This work benefited from discussions with Frank Keller.
This work was supported by the Qatar Computing Research Institute (QCRI) and Samsung Research.  

\bibliographystyle{iclr2018_conference}
\bibliography{references}

\begin{thebibliography}{55}
\providecommand{\natexlab}[1]{#1}
\providecommand{\url}[1]{\texttt{#1}}
\expandafter\ifx\csname urlstyle\endcsname\relax
  \providecommand{\doi}[1]{doi: #1}\else
  \providecommand{\doi}{doi: \begingroup \urlstyle{rm}\Url}\fi

\bibitem[Bahdanau et~al.(2014)Bahdanau, Cho, and Bengio]{bahdanau2014neural}
Dzmitry Bahdanau, Kyunghyun Cho, and Yoshua Bengio.
\newblock {Neural Machine Translation by Jointly Learning to Align and
  Translate}.
\newblock \emph{arXiv preprint arXiv:1409.0473}, 2014.

\bibitem[Belinkov \& Glass(2016)Belinkov and Glass]{belinkov-glass:2016:SeMaT}
Yonatan Belinkov and James Glass.
\newblock {Large-Scale Machine Translation between Arabic and Hebrew: Available
  Corpora and Initial Results}.
\newblock In \emph{Proceedings of the Workshop on Semitic Machine Translation},
  pp.\  7--12, Austin, Texas, November 2016. Association for Computational
  Linguistics.

\bibitem[Belinkov et~al.(2017{\natexlab{a}})Belinkov, Durrani, Dalvi, Sajjad,
  and Glass]{belinkov-EtAl:2017:Long}
Yonatan Belinkov, Nadir Durrani, Fahim Dalvi, Hassan Sajjad, and James Glass.
\newblock {What do Neural Machine Translation Models Learn about Morphology?}
\newblock In \emph{Proceedings of the 55th Annual Meeting of the Association
  for Computational Linguistics (Volume 1: Long Papers)}, pp.\  861--872,
  Vancouver, Canada, July 2017{\natexlab{a}}.

\bibitem[Belinkov et~al.(2017{\natexlab{b}})Belinkov, M\`arquez, Sajjad, Dalvi,
  Durrani, and Glass]{belinkov:2017:IJCNLP}
Yonatan Belinkov, Llu\'{i}s M\`arquez, Hassan Sajjad, Fahim Dalvi, Nadir
  Durrani, and James Glass.
\newblock {Evaluating Layers of Representation in Neural Machine Translation on
  Part-of-Speech and Semantic Tagging Tasks}.
\newblock In \emph{Proceedings of the 8th International Joint Conference on
  Natural Language Processing (Volume 1: Long Papers)}, Taipei, Taiwan,
  November 2017{\natexlab{b}}. Association for Computational Linguistics.

\bibitem[Biggio et~al.(2012)Biggio, Nelson, and
  Laskov]{Biggio:2012:PAA:3042573.3042761}
Battista Biggio, Blaine Nelson, and Pavel Laskov.
\newblock {Poisoning Attacks Against Support Vector Machines}.
\newblock In \emph{Proceedings of the 29th International Coference on
  International Conference on Machine Learning}, ICML'12, pp.\  1467--1474,
  USA, 2012. Omnipress.
\newblock ISBN 978-1-4503-1285-1.
\newblock URL \url{http://dl.acm.org/citation.cfm?id=3042573.3042761}.

\bibitem[Bishop(1995)]{Bishop:1995}
Christopher Bishop.
\newblock {Training with noise is equivalent to Tikhonov regularization}.
\newblock \emph{Neural Computation}, 7:\penalty0 108--116, January 1995.

\bibitem[Cettolo et~al.(2012)Cettolo, Girardi, and
  Federico]{cettoloEtAl:EAMT2012}
Mauro Cettolo, Christian Girardi, and Marcello Federico.
\newblock {WIT$^3$: Web Inventory of Transcribed and Translated Talks}.
\newblock In \emph{Proceedings of the 16$^{th}$ Conference of the European
  Association for Machine Translation (EAMT)}, pp.\  261--268, Trento, Italy,
  May 2012.

\bibitem[Chan et~al.(2017)Chan, He, Li, and Hsu]{Chan2017}
Patrick P.~K. Chan, Zhi-Min He, Hongjiang Li, and Chien-Chang Hsu.
\newblock Data sanitization against adversarial label contamination based on
  data complexity.
\newblock \emph{International Journal of Machine Learning and Cybernetics}, Jan
  2017.
\newblock ISSN 1868-808X.
\newblock \doi{10.1007/s13042-016-0629-5}.
\newblock URL \url{https://doi.org/10.1007/s13042-016-0629-5}.

\bibitem[Costa-juss\`{a} \& Fonollosa(2016)Costa-juss\`{a} and
  Fonollosa]{costajussa-fonollosa:2016:P16-2}
Marta~R. Costa-juss\`{a} and Jos\'{e} A.~R. Fonollosa.
\newblock {Character-based Neural Machine Translation}.
\newblock In \emph{Proceedings of the 54th Annual Meeting of the Association
  for Computational Linguistics (Volume 2: Short Papers)}, pp.\  357--361,
  Berlin, Germany, August 2016. Association for Computational Linguistics.
\newblock URL \url{http://anthology.aclweb.org/P16-2058}.

\bibitem[Cretu et~al.(2008)Cretu, Stavrou, Locasto, Stolfo, and
  Keromytis]{Cretu:2008:COD:1397759.1398062}
Gabriela~F. Cretu, Angelos Stavrou, Michael~E. Locasto, Salvatore~J. Stolfo,
  and Angelos~D. Keromytis.
\newblock {Casting out Demons: Sanitizing Training Data for Anomaly Sensors}.
\newblock In \emph{Proceedings of the 2008 IEEE Symposium on Security and
  Privacy}, SP '08, pp.\  81--95, Washington, DC, USA, 2008. IEEE Computer
  Society.
\newblock ISBN 978-0-7695-3168-7.
\newblock \doi{10.1109/SP.2008.11}.
\newblock URL \url{https://doi.org/10.1109/SP.2008.11}.

\bibitem[Dalvi et~al.(2017)Dalvi, Durrani, Sajjad, Belinkov, and
  Vogel]{dalvi:2017:IJCNLP}
Fahim Dalvi, Nadir Durrani, Hassan Sajjad, Yonatan Belinkov, and Stephan Vogel.
\newblock {Understanding and Improving Morphological Learning in the Neural
  Machine Translation Decoder}.
\newblock In \emph{Proceedings of the 8th International Joint Conference on
  Natural Language Processing (Volume 1: Long Papers)}, Taipei, Taiwan,
  November 2017. Association for Computational Linguistics.

\bibitem[Ebrahimi et~al.(2017)Ebrahimi, Rao, Lowd, and
  Dou]{ebrahimi2017hotflip}
Javid Ebrahimi, Anyi Rao, Daniel Lowd, and Dejing Dou.
\newblock {HotFlip: White-Box Adversarial Examples for NLP}.
\newblock \emph{arXiv preprint arXiv:1712.06751}, 2017.

\bibitem[Gao et~al.(2018)Gao, Lanchantin, Soffa, and Qi]{gao2018black}
Ji~Gao, Jack Lanchantin, Mary~Lou Soffa, and Yanjun Qi.
\newblock {Black-box Generation of Adversarial Text Sequences to Evade Deep
  Learning Classifiers}.
\newblock \emph{arXiv preprint arXiv:1801.04354}, 2018.

\bibitem[Globerson \& Roweis(2006)Globerson and
  Roweis]{Globerson:2006:NTT:1143844.1143889}
Amir Globerson and Sam Roweis.
\newblock {Nightmare at Test Time: Robust Learning by Feature Deletion}.
\newblock In \emph{Proceedings of the 23rd International Conference on Machine
  Learning}, ICML '06, pp.\  353--360, New York, NY, USA, 2006. ACM.
\newblock ISBN 1-59593-383-2.
\newblock \doi{10.1145/1143844.1143889}.
\newblock URL \url{http://doi.acm.org/10.1145/1143844.1143889}.

\bibitem[Goodfellow et~al.(2015)Goodfellow, Shlens, and
  Szegedy]{goodfellow2014explaining}
Ian~J Goodfellow, Jonathon Shlens, and Christian Szegedy.
\newblock {Explaining and Harnessing Adversarial Examples}.
\newblock In \emph{International Conference on Learning Representations
  (ICLR)}, 2015.

\bibitem[Heigold et~al.(2017)Heigold, Neumann, and van
  Genabith]{heigold2017robust}
Georg Heigold, G{\"u}nter Neumann, and Josef van Genabith.
\newblock {How Robust Are Character-Based Word Embeddings in Tagging and MT
  Against Wrod Scramlbing or Randdm Nouse?}
\newblock \emph{arXiv preprint arXiv:1704.04441}, 2017.

\bibitem[Hochreiter \& Schmidhuber(1997)Hochreiter and
  Schmidhuber]{hochreiter1997long}
Sepp Hochreiter and J{\"u}rgen Schmidhuber.
\newblock {Long short-term memory}.
\newblock \emph{Neural Computation}, 9\penalty0 (8):\penalty0 1735--1780, 1997.

\bibitem[Jia \& Liang(2017)Jia and Liang]{jia-liang:2017:EMNLP2017}
Robin Jia and Percy Liang.
\newblock {Adversarial Examples for Evaluating Reading Comprehension Systems}.
\newblock In \emph{Proceedings of the 2017 Conference on Empirical Methods in
  Natural Language Processing}, pp.\  2011--2021, Copenhagen, Denmark,
  September 2017.

\bibitem[Junczys-Dowmunt \& Birch(2016)Junczys-Dowmunt and
  Birch]{junczys2016university}
Marcin Junczys-Dowmunt and Alexandra Birch.
\newblock {The University of Edinburgh's systems submission to the MT task at
  IWSLT}.
\newblock In \emph{Proceedings of the First Conference on Machine Translation},
  Seattle, USA, 2016.

\bibitem[Kim(2016)]{kim2016}
Yoon Kim.
\newblock Seq2seq-attn.
\newblock \url{https://github.com/harvardnlp/seq2seq-attn}, 2016.

\bibitem[Kim et~al.(2015)Kim, Jernite, Sontag, and Rush]{kim2015character}
Yoon Kim, Yacine Jernite, David Sontag, and Alexander~M Rush.
\newblock {Character-aware Neural Language Models}.
\newblock \emph{arXiv preprint arXiv:1508.06615}, 2015.

\bibitem[Lee et~al.(2017)Lee, Cho, and Hofmann]{lee2017}
Jason Lee, Kyunghyun Cho, and Thomas Hofmann.
\newblock {Fully Character-Level Neural Machine Translation without Explicit
  Segmentation}.
\newblock \emph{Transactions of the Association for Computational Linguistics
  (TACL)}, 2017.

\bibitem[Liang et~al.(2017)Liang, Li, Su, Bian, Li, and Shi]{liang2017deep}
Bin Liang, Hongcheng Li, Miaoqiang Su, Pan Bian, Xirong Li, and Wenchang Shi.
\newblock {Deep Text Classification Can be Fooled}.
\newblock \emph{arXiv preprint arXiv:1704.08006}, 2017.

\bibitem[Liu et~al.(2017)Liu, Chen, Liu, and Song]{liu2016delving}
Yanpei Liu, Xinyun Chen, Chang Liu, and Dawn Song.
\newblock {Delving into Transferable Adversarial Examples and Black-box
  Attacks}.
\newblock 2017.

\bibitem[Matsuoka(1992)]{Matsuoka:1992}
K.~Matsuoka.
\newblock Noise injection into inputs in back-propagation learning.
\newblock \emph{IEEE Transactions on Systems, Man, and Cybernetics},
  22\penalty0 (3):\penalty0 436--440, May 1992.

\bibitem[Max \& Wisniewski(2010)Max and Wisniewski]{max10wicopaco}
Aurélien Max and Guillaume Wisniewski.
\newblock {Mining Naturally-occurring Corrections and Paraphrases from
  Wikipedia’s Revision History}.
\newblock In \emph{Proceedings of the Seventh conference on International
  Language Resources and Evaluation (LREC'10)}, Valletta, Malta, may 2010.
  European Language Resources Association (ELRA).
\newblock ISBN 2-9517408-6-7.
\newblock URL \url{https://wicopaco.limsi.fr}.

\bibitem[Mayall et~al.(1997)Mayall, Humphreys, and Olson]{Mayall}
K.~Mayall, G.W. Humphreys, and A.~Olson.
\newblock {Disruption to word or letter processing? The origins of case-mixing
  effects}.
\newblock \emph{Journal of Experimental Psychology: Learning, Memory, \&
  Cognition}, 23:\penalty0 1275 -- 1286, 1997.

\bibitem[McCusker et~al.(1981)McCusker, Gough, and Bias]{swapping}
L.~X. McCusker, P.~B. Gough, and R.~G. Bias.
\newblock Word recognition inside out and outside in.
\newblock \emph{Journal of Experimental Psychology: Human Perception and
  Performance}, 7\penalty0 (3):\penalty0 538 -- 551, 1981.

\bibitem[Mei \& Zhu(2015)Mei and Zhu]{Mei:2015:UMT:2886521.2886721}
Shike Mei and Xiaojin Zhu.
\newblock {Using Machine Teaching to Identify Optimal Training-set Attacks on
  Machine Learners}.
\newblock In \emph{Proceedings of the Twenty-Ninth AAAI Conference on
  Artificial Intelligence}, AAAI'15, pp.\  2871--2877. AAAI Press, 2015.
\newblock ISBN 0-262-51129-0.
\newblock URL \url{http://dl.acm.org/citation.cfm?id=2886521.2886721}.

\bibitem[Narodytska \& Kasiviswanathan(2017)Narodytska and
  Kasiviswanathan]{8014906}
N.~Narodytska and S.~Kasiviswanathan.
\newblock {Simple Black-Box Adversarial Attacks on Deep Neural Networks}.
\newblock In \emph{2017 IEEE Conference on Computer Vision and Pattern
  Recognition Workshops (CVPRW)}, pp.\  1310--1318, July 2017.
\newblock \doi{10.1109/CVPRW.2017.172}.

\bibitem[Papernot et~al.(2016{\natexlab{a}})Papernot, McDaniel, and
  Goodfellow]{papernot2016transferability}
Nicolas Papernot, Patrick McDaniel, and Ian Goodfellow.
\newblock {Transferability in Machine Learning: from Phenomena to Black-Box
  Attacks using Adversarial Samples}.
\newblock \emph{arXiv preprint arXiv:1605.07277}, 2016{\natexlab{a}}.

\bibitem[Papernot et~al.(2016{\natexlab{b}})Papernot, McDaniel, Swami, and
  Harang]{papernot2016crafting}
Nicolas Papernot, Patrick McDaniel, Ananthram Swami, and Richard Harang.
\newblock {Crafting Adversarial Input Sequences for Recurrent Neural Networks}.
\newblock In \emph{Military Communications Conference, MILCOM 2016-2016 IEEE},
  pp.\  49--54. IEEE, 2016{\natexlab{b}}.

\bibitem[Papernot et~al.(2017)Papernot, McDaniel, Goodfellow, Jha, Celik, and
  Swami]{Papernot:2017:PBA:3052973.3053009}
Nicolas Papernot, Patrick McDaniel, Ian Goodfellow, Somesh Jha, Z.~Berkay
  Celik, and Ananthram Swami.
\newblock {Practical Black-Box Attacks Against Machine Learning}.
\newblock In \emph{Proceedings of the 2017 ACM on Asia Conference on Computer
  and Communications Security}, ASIA CCS '17, pp.\  506--519, New York, NY,
  USA, 2017. ACM.
\newblock ISBN 978-1-4503-4944-4.
\newblock \doi{10.1145/3052973.3053009}.
\newblock URL \url{http://doi.acm.org/10.1145/3052973.3053009}.

\bibitem[Pelli et~al.(2003)Pelli, Farell, and Moore]{Pelli}
D.~G. Pelli, B.~Farell, and D.C. Moore.
\newblock The remarkable inefficiency of word recognition.
\newblock \emph{Nature}, 423:\penalty0 752 -- 756, 2003.

\bibitem[Rawlinson(1976)]{rawlinson}
G.~E. Rawlinson.
\newblock \emph{The significance of letter position in word recognition}.
\newblock PhD thesis, 1976.

\bibitem[Rayner et~al.(2006)Rayner, White, Johnson, and Liversedge]{slower}
Keith Rayner, Sarah~J. White, Rebecca~L. Johnson, and Simon~P. Liversedge.
\newblock {Raeding Wrods With Jubmled Lettres: There Is a Cost}.
\newblock \emph{Psychological Science}, 17\penalty0 (3):\penalty0 192 -- 193,
  2006.

\bibitem[Reicher(1969)]{Reicher}
G.~M. Reicher.
\newblock Perceptual recognition as a function of meaningfulness of stimulus
  material.
\newblock \emph{Journal of Experimental Psychology}, 81\penalty0 (2):\penalty0
  275 -- 280, 1969.

\bibitem[Rubinstein et~al.(2009)Rubinstein, Nelson, Huang, Joseph, Lau, Rao,
  Taft, and Tygar]{Rubinstein:2009:AUD:1644893.1644895}
Benjamin~I.P. Rubinstein, Blaine Nelson, Ling Huang, Anthony~D. Joseph,
  Shing-hon Lau, Satish Rao, Nina Taft, and J.~D. Tygar.
\newblock {ANTIDOTE: Understanding and Defending Against Poisoning of Anomaly
  Detectors}.
\newblock In \emph{Proceedings of the 9th ACM SIGCOMM Conference on Internet
  Measurement}, IMC '09, pp.\  1--14, New York, NY, USA, 2009. ACM.
\newblock ISBN 978-1-60558-771-4.
\newblock \doi{10.1145/1644893.1644895}.
\newblock URL \url{http://doi.acm.org/10.1145/1644893.1644895}.

\bibitem[Saberi \& Perrott(1999)Saberi and Perrott]{saberi}
Kourosh Saberi and David~R. Perrott.
\newblock Cognitive restoration of reversed speech.
\newblock \emph{Nature}, 398\penalty0 (760), April 1999.

\bibitem[Sajjad et~al.(2017)Sajjad, Dalvi, Durrani, Abdelali, Belinkov, and
  Vogel]{sajjad:2017:ACL}
Hassan Sajjad, Fahim Dalvi, Nadir Durrani, Ahmed Abdelali, Yonatan Belinkov,
  and Stephan Vogel.
\newblock {Challenging Language-Dependent Segmentation for Arabic: An
  Application to Machine Translation and Part-of-Speech Tagging}.
\newblock In \emph{Proceedings of the 55th Annual Meeting of the Association
  for Computational Linguistics}, Vancouver, Canada, July 2017. Association for
  Computational Linguistics.

\bibitem[Sakaguchi et~al.(2017)Sakaguchi, Duh, Post, and
  Durme]{DBLP:conf/aaai/SakaguchiDPD17}
Keisuke Sakaguchi, Kevin Duh, Matt Post, and Benjamin~Van Durme.
\newblock {Robsut Wrod Reocginiton via Semi-Character Recurrent Neural
  Network}.
\newblock In \emph{Proceedings of the Thirty-First {AAAI} Conference on
  Artificial Intelligence, February 4-9, 2017, San Francisco, California,
  {USA.}}, pp.\  3281--3287. {AAAI} Press, 2017.
\newblock URL \url{http://aaai.org/ocs/index.php/AAAI/AAAI17/paper/view/14332}.

\bibitem[Samanta \& Mehta(2017)Samanta and Mehta]{samanta2017towards}
Suranjana Samanta and Sameep Mehta.
\newblock {Towards Crafting Text Adversarial Samples}.
\newblock \emph{arXiv preprint arXiv:1707.02812}, 2017.

\bibitem[{\v S}ebesta et~al.(2017){\v S}ebesta, Bed{\v r}ichov{\'a}, {\v
  S}ormov{\'a}, {\v S}tindlov{\'a}, Hrdli{\v c}ka, Hrdli{\v c}kov{\'a}, Hana,
  Petkevi{\v c}, Jel{\'i}nek, {\v S}kodov{\'a}, Jane{\v s}, Lund{\'a}kov{\'a},
  Skoumalov{\'a}, Sl{\'a}dek, Pierscieniak, Toufarov{\'a}, Straka, Rosen,
  N{\'a}plava, and Pol{\'a}{\v c}kov{\'a}]{CzeSL}
Karel {\v S}ebesta, Zuzanna Bed{\v r}ichov{\'a}, Kate{\v r}ina {\v
  S}ormov{\'a}, Barbora {\v S}tindlov{\'a}, Milan Hrdli{\v c}ka, Tereza
  Hrdli{\v c}kov{\'a}, Ji{\v r}{\'i} Hana, Vladim{\'i}r Petkevi{\v c},
  Tom{\'a}{\v s} Jel{\'i}nek, Svatava {\v S}kodov{\'a}, Petr Jane{\v s},
  Kate{\v r}ina Lund{\'a}kov{\'a}, Hana Skoumalov{\'a}, {\v S}imon Sl{\'a}dek,
  Piotr Pierscieniak, Dagmar Toufarov{\'a}, Milan Straka, Alexandr Rosen, Jakub
  N{\'a}plava, and Marie Pol{\'a}{\v c}kov{\'a}.
\newblock {CzeSL} grammatical error correction dataset ({CzeSL}-{GEC}).
\newblock Technical report, {LINDAT}/{CLARIN} digital library at the Institute
  of Formal and Applied Linguistics, Charles University, 2017.
\newblock URL
  \url{https://lindat.mff.cuni.cz/repository/xmlui/handle/11234/1-2143}.

\bibitem[Sennrich(2017)]{E17-2060}
Rico Sennrich.
\newblock {How Grammatical is Character-level Neural Machine Translation?
  Assessing MT Quality with Contrastive Translation Pairs}.
\newblock In \emph{Proceedings of the 15th Conference of the European Chapter
  of the Association for Computational Linguistics: Volume 2, Short Papers},
  pp.\  376--382. Association for Computational Linguistics, 2017.
\newblock URL \url{http://aclweb.org/anthology/E17-2060}.

\bibitem[Sennrich et~al.(2016{\natexlab{a}})Sennrich, Haddow, and
  Birch]{P16-1162}
Rico Sennrich, Barry Haddow, and Alexandra Birch.
\newblock {Neural Machine Translation of Rare Words with Subword Units}.
\newblock In \emph{Proceedings of the 54th Annual Meeting of the Association
  for Computational Linguistics (Volume 1: Long Papers)}, pp.\  1715--1725.
  Association for Computational Linguistics, 2016{\natexlab{a}}.
\newblock \doi{10.18653/v1/P16-1162}.
\newblock URL
  \url{http://aclanthology.coli.uni-saarland.de/pdf/P/P16/P16-1162.pdf}.

\bibitem[Sennrich et~al.(2016{\natexlab{b}})Sennrich, Haddow, and
  Birch]{sennrich-haddow-birch:2016:WMT}
Rico Sennrich, Barry Haddow, and Alexandra Birch.
\newblock {Edinburgh Neural Machine Translation Systems for WMT 16}.
\newblock In \emph{Proceedings of the First Conference on Machine Translation},
  pp.\  371--376, Berlin, Germany, August 2016{\natexlab{b}}. Association for
  Computational Linguistics.

\bibitem[Sennrich et~al.(2017)Sennrich, Firat, Cho, Birch, Haddow, Hitschler,
  Junczys-Dowmunt, L\"{a}ubli, Miceli~Barone, Mokry, and Nadejde]{sennrich2017}
Rico Sennrich, Orhan Firat, Kyunghyun Cho, Alexandra Birch, Barry Haddow,
  Julian Hitschler, Marcin Junczys-Dowmunt, Samuel L\"{a}ubli, Antonio~Valerio
  Miceli~Barone, Jozef Mokry, and Maria Nadejde.
\newblock {Nematus: a Toolkit for Neural Machine Translation}.
\newblock In \emph{Proceedings of the Software Demonstrations of the 15th
  Conference of the European Chapter of the Association for Computational
  Linguistics}, pp.\  65--68, Valencia, Spain, April 2017. Association for
  Computational Linguistics.
\newblock URL \url{http://aclweb.org/anthology/E17-3017}.

\bibitem[Shi et~al.(2016)Shi, Padhi, and
  Knight]{shi-padhi-knight:2016:EMNLP2016}
Xing Shi, Inkit Padhi, and Kevin Knight.
\newblock {Does String-Based Neural MT Learn Source Syntax?}
\newblock In \emph{Proceedings of the 2016 Conference on Empirical Methods in
  Natural Language Processing}, pp.\  1526--1534, Austin, Texas, November 2016.
  Association for Computational Linguistics.
\newblock URL \url{https://aclweb.org/anthology/D16-1159}.

\bibitem[Sutskever et~al.(2014)Sutskever, Vinyals, and
  Le]{sutskever2014sequence}
Ilya Sutskever, Oriol Vinyals, and Quoc~VV Le.
\newblock {Sequence to Sequence Learning with Neural Networks}.
\newblock In \emph{Advances in neural information processing systems}, pp.\
  3104--3112, 2014.

\bibitem[Szegedy et~al.(2014)Szegedy, Zaremba, Sutskever, Bruna, Erhan,
  Goodfellow, and Fergus]{szegedy2013intriguing}
Christian Szegedy, Wojciech Zaremba, Ilya Sutskever, Joan Bruna, Dumitru Erhan,
  Ian Goodfellow, and Rob Fergus.
\newblock Intriguing properties of neural networks.
\newblock In \emph{International Conference on Learning Representations
  (ICLR)}, 2014.

\bibitem[Tram{\`e}r et~al.(2017)Tram{\`e}r, Kurakin, Papernot, Boneh, and
  McDaniel]{tramer2017ensemble}
Florian Tram{\`e}r, Alexey Kurakin, Nicolas Papernot, Dan Boneh, and Patrick
  McDaniel.
\newblock {Ensemble Adversarial Training: Attacks and Defenses}.
\newblock \emph{arXiv preprint arXiv:1705.07204}, 2017.

\bibitem[Vylomova et~al.(2016)Vylomova, Cohn, He, and
  Haffari]{vylomova2016word}
Ekaterina Vylomova, Trevor Cohn, Xuanli He, and Gholamreza Haffari.
\newblock {Word Representation Models for Morphologically Rich Languages in
  Neural Machine Translation}.
\newblock \emph{arXiv preprint arXiv:1606.04217}, 2016.

\bibitem[Wisniewski et~al.(2013)Wisniewski, Schöne, Nicolas, Vettori, Boyd,
  Meurers, Abel, and Hana]{MERLIN}
Katrin Wisniewski, Karin Schöne, Lionel Nicolas, Chiara Vettori, Adriane Boyd,
  Detmar Meurers, Andrea Abel, and Jirka Hana.
\newblock {MERLIN: An online trilingual learner corpus empirically grounding
  the European Reference Levels in authentic learner data}, 10 2013.
\newblock URL
  \url{https://www.ukp.tu-darmstadt.de/data/spelling-correction/rwse-datasets}.

\bibitem[Wu et~al.(2016)Wu, Schuster, Chen, Le, Norouzi, Macherey, Krikun, Cao,
  Gao, Macherey, et~al.]{wu2016google}
Yonghui Wu, Mike Schuster, Zhifeng Chen, Quoc~V Le, Mohammad Norouzi, Wolfgang
  Macherey, Maxim Krikun, Yuan Cao, Qin Gao, Klaus Macherey, et~al.
\newblock Google's neural machine translation system: Bridging the gap between
  human and machine translation.
\newblock \emph{arXiv preprint arXiv:1609.08144}, 2016.

\bibitem[Zesch(2012)]{zesch:2012:EACL2012}
Torsten Zesch.
\newblock {Measuring Contextual Fitness Using Error Contexts Extracted from the
  Wikipedia Revision History}.
\newblock In \emph{Proceedings of the 13th Conference of the European Chapter
  of the Association for Computational Linguistics}, pp.\  529--538, Avignon,
  France, April 2012. Association for Computational Linguistics.

\end{thebibliography}

\end{document}